\documentclass{article}
\usepackage{stywhispers,amsmath,epsfig}
\usepackage{bbm}
\DeclareMathOperator*{\argmin}{argmin}
\usepackage{algorithm}
\usepackage{algpseudocode}
\title{Unbalanced Optimal Transport Dictionary Learning for Unsupervised Hyperspectral Image Clustering}
\name{Joshua Lentz$^{1}$, Nicholas Karris$^{2}$, Alex Cloninger$^{2, 3}$, James M. Murphy$^{1}$\thanks{Joshua Lentz and James M. Murphy are supported by NSF grant DMS-2318894 and Alex Cloninger is supported by NSF grant CISE-2403452.}}
\address{$^{1}$Tufts University, Department of Mathematics, Medford, MA 02155, USA \\ $^{2}$University of California San Diego, Department of Mathematics, La Jolla, CA 92093, USA \\ $^{3}$University of California San Diego, Halıcıoğlu Data Science Institute, La Jolla, CA 92093, USA
}

\begin{document}
%\ninept
%
\maketitle
\author{Joshua Lentz$^{1}$, Nicholas Karris$^{2}$, Alex Cloninger$^{2}$, James Murphy$^{1}$ \\
$^{1}$Tufts University, Department of Mathematics \\ Medford, MA 02155, USA\\
$^{2}$
}
\begin{abstract}
Hyperspectral images capture vast amounts of high-dimensional spectral information about a scene, making labeling an intensive task that is resistant to out-of-the-box statistical methods. Unsupervised learning of clusters allows for automated segmentation of the scene, enabling a more rapid understanding of the image. Partitioning the spectral information contained within the data via dictionary learning in Wasserstein space has proven an effective method for unsupervised clustering. However, this approach requires balancing the spectral profiles of the data, blurring the classes, and sacrificing robustness to outliers and noise. In this paper, we suggest improving this approach by utilizing unbalanced Wasserstein barycenters to learn a lower-dimensional representation of the underlying data. The deployment of spectral clustering on the learned representation results in an effective approach for the unsupervised learning of labels. 
\end{abstract}
\section{Introduction}
\label{sec:intro}
Hyperspectral imaging captures a huge amount of spatial and spectral data from materially diverse scenes. Effectively labeling the resulting data set typically requires large quantities of hand-labeled training data, which are often difficult and time-consuming to acquire. In the case that such labeled data are available, supervised learning methods such as support vector machines \cite{melgani2004classifiation}, random forests \cite{ham2005investigation}, and deep learning approaches \cite{li2019deep, kumar2024deep} provide high-accuracy labels of HSI data \cite{ahmad2025comprehensive}. The lack of labeled training data has motivated the development of semi-supervised and unsupervised learning methods \cite{zhai2021hyperspectral} which require little or no training data.  Of particular challenge is the high dimensionality of the underlying data, which makes standard statistical learning techniques unreliable. \\
\indent In this paper, we seek to further develop a prior unsupervised approach that deployed dictionary learning in the Wasserstein space. The core idea of this method is to represent each HSI data point as a probability distribution supported on the space of its reflectances. We then deploy iterative methods to learn a set of dictionary atoms and weights which reconstruct the original data via the (nonlinear) Wasserstein barycenter. This set of weights provides a lower-dimensional representation of the original data on which spectral clustering was performed, providing unsupervised labeling of HSI scenes. One drawback of this approach was the necessity to normalize the HSI data as probability distributions, potentially blurring classes, and sacrificing robustness to outliers and noisy data. As an improvement, we suggest modifying the prior approach to instead use \emph{unbalanced Wasserstein barycenters}, allowing for the differing total masses of the HSI data points. Moreover, unbalanced optimal transport has been shown to be more robust to outliers in statistical applications \cite{balaji2020robust}. We refer to this approach as Unbalanced Optimal Transport Dictionary Learning, and explore its implementation and improvements below. \\
\indent The rest of this paper is organized as follows: in Section 2, we review optimal transport, unbalanced optimal transport, and Wasserstein dictionary learning as it relates to HSI labeling. In Section 3, we discuss our modification to Wasserstein dictionary learning, both in theory and in practice. In Section 4, we provide numerical results on a suite of known HSI labeling benchmarks. We conclude in Section 5 by providing directions for future research and other areas of interest which may benefit from our method. 
%Im not sure what these other areas are but it sounded good. 

\section{BACKGROUND}
\label{sec:format}

Unsupervised learning analyzes HSI data, which we view as distributions, $\{\mu_i\}_{i=1}^n \in \mathbbm{R}^d$, and generates a label set $\{y_i\}_{i=1}^n$ with each $y_i \in \{1,2,..,K\}$ where $K$ is the number of clusters. The value for $K$ can either be learned or estimated, though a clean measurement of labeling accuracy only exists when our chosen number of clusters matches the number of ground truth classes. Typical methods seek to find geometric and statistical patterns in the dataset $\{\mu_i\}_{i=1}^n$ to infer clusters \cite{shi2000normalized}. We seek to build on a class of unsupervised approaches which leverage spectral clustering \cite{ng2001spectral}, which parses the structure of a latent graph to cluster the underlying data \cite{cahill2014schrodinger}. For some metric $d$,  we define $W \in \mathbbm{R}^{n \times n}$ to be the $k$-nearest neighbors graph with respect to some metric for suitable $k$. Clustering is then performed by analyzing the lowest frequency eigenvectors of the normalized graph Laplacian $L = I - D^{-\frac{1}{2}}WD^{-\frac{1}{2}}$ where $D \in \mathbbm{R}^{n \times n}$ is the diagonal degree matrix with $D_{ii} = \sum_{j = 1}^n W_{ij}$ for all $i = 1, ..., n$ \cite{vonLuxburg2007tutorial}. Typically, we then perform $k$-means on the $k$ eigenvectors of $L$ with the smallest eigenvalues. 
\\
\indent Previously, \cite{fullenbaum2024hyperspectral, fullenbaum2024nonlinear}  considered recasting HSI pixels as probability distributions. They then leveraged iterative methods to find a learned representation of the underlying data via the nonlinear reconstruction of the Wasserstein barycenter. This method was the so-called Wasserstein dictionary learning \cite{fullenbaum2024hyperspectral, schmitz2018wasserstein, mueller2023geometrically}.  

%TODO
%Graphics?

\indent Optimal transport, and in particular the Wasserstein distance, is naturally suited for comparing probability distributions. For the sake of simplicity, we will introduce the definitions as they relate to 1-dimensional discrete optimal transport. For a deeper exploration of optimal transport, we refer the reader to \cite{villani2008optimal}. In this narrow case, we restrict ourselves to probability vectors $\mu = (\mu_1,...,\mu_n)$ and $\nu = (\nu_1,..,\nu_m)$. Define the set of couplings $\Pi(\mu, \nu) \subset \mathbbm{R}_{\geq 0}^{n \times m}$ as all $X$ such  that $\sum_{j=1}^m X_{ij} = \mu_i$ and $\sum_{i=1}^n X_{ij} = \nu_j$. Let $C \in \mathbbm{R}_{\geq 0}^{n \times m}$ be a cost matrix given by $C_{ij} = |x_i - y_j|^2$ where $x_i, y_j$ are in the support of $\mu$ and $\nu$ respectively. In order to guarantee strict convexity and efficient computation, it is standard to add an entropy term, $\epsilon H(\pi) = \epsilon \sum_{i=1}^n\sum_{j=1}^m \pi_{ij}\log(\pi_{ij})$,  to the problem, which is known to give accurate approximations of the true solution \cite{cuturi2013sinkhorn}. Lastly, define $\langle A, B \rangle = \sum_{i=1}^n \sum_{j=1}^m A_{ij}B_{ij}$ to be the Frobenius inner product. Then the entropic regularized optimal transport problem seeks to minimize (1) for some $\epsilon>0$. The effect of $\epsilon$ is to act as a smoothing factor on the transport plan, with larger $\epsilon$ blurring the solution to the unregularized problem. 
\begin{equation}
\begin{aligned}
    OT_\epsilon(\mu, \nu) = \min_{X \in \Pi(\mu, \nu)} \langle C, X \rangle + \epsilon H(X).
\end{aligned}
\end{equation}
\indent We call the above transport problem balanced due to the strict requirement that both distributions, $\mu$ and $\nu$, have the same total mass. %This allows a one-to-one matching of mass when constructing a transport plan. 
Unlike optimal transport, unbalanced optimal transport seeks to lift this restriction and instead penalizes the marginal terms, $X_1, X_2$, using Csisz\'{a}r divergences; in this case, we use the KL divergence \cite{chizat2015unbalanced}. In unbalanced optimal transport, transport still occurs, but some mass is created/destroyed as transport takes place. As before, we incorporate an entropic regularization term enabling the use of a Sinkhorn-like algorithm for more efficient computation \cite{chizat2017scaling}. The entropic unbalanced optimal transport problem (in one dimension on discretely supported measures) is as follows: given vectors $\mu = (\mu_1, .., \mu_n)$ and $\nu = (\nu_1, ..., \nu_m)$ such that $\mu_i, \nu_i \geq 0$ find $X \in \mathbbm{R}_{\geq 0}^{n \times m}$ minimizing the following:
\begin{equation}
\begin{aligned}
    UOT_\epsilon^\tau(\mu, \nu) = &\min_{X \in \Pi(\mu, \nu)} \langle X, C\rangle + \tau KL(X \mathbbm{1}_m || \mu)  \\ 
    &+ \tau KL(X^T \mathbbm{1}_n || \nu) + \epsilon KL( X || \mu\nu^T).
\end{aligned}
\end{equation}
Here, $\tau$  and $ \epsilon$ are the marginal and entropic regularization terms respectively, $\mathbbm{1}_n$ is the vector of $n$ 1's, and $\mu\nu^T \in \mathbbm{R}^{n \times m}$ is the outer product of $\mu$ and $\nu$. The difference in entropy term when compared to (1) arises from the fact that $\int d\mu d\nu \ne 1$. Indeed, the entropy term, $H(X)$, differs from KL divergence by a constant when $\int d\mu d\nu \ne 1$. 
Moreover, this entropy term also melds nicely with a Sinkhorn-like algorithm, see \cite{chizat2017scaling}. 

\indent A natural question in the context of optimal transport is to ask for an average of distributions, probabilities or otherwise. This leads us to the concept of the barycenter, which is the distribution minimizing the transport cost from a weighted set of distributions onto itself. For a set of $m$ distributions $\{\nu_j\}_{j=1}^m \in \mathbbm{R}^d_{\geq 0}$ and a vector of weights $\Lambda = (\lambda_1, ..., \lambda_m)$ such that $\sum_{i=1}^m \lambda_i = 1$, we can define the entropic unbalanced optimal transport barycenter \cite{chizat2017scaling, agueh2011barycenters}   to be: 
\begin{equation}
\begin{aligned}
%Unclear how we want to notate this. I have P in my documents because I was borrowing the notation used in the WDL paper, but this is less standard. We should also clarify this definition is notated as if the underlying distribibutions have the same support. 
        P( \{ \nu_j\}_{j=1}^m, \Lambda) = \argmin_{\mu \in \mathbbm{R}^d_{\geq 0}} \sum_{j=1}^m \lambda_i UOT_\epsilon^\tau(\mu, \nu_j).
\end{aligned}
\end{equation}
For an excellent exploration of balanced barycenters in the space of probability distributions, we refer the reader to \cite{agueh2011barycenters}. For the first treatment of unbalanced barycenters, see \cite{chizat2017scaling}. 
\\
\begin{figure}
    \centering
    \includegraphics[width=1.1\linewidth]{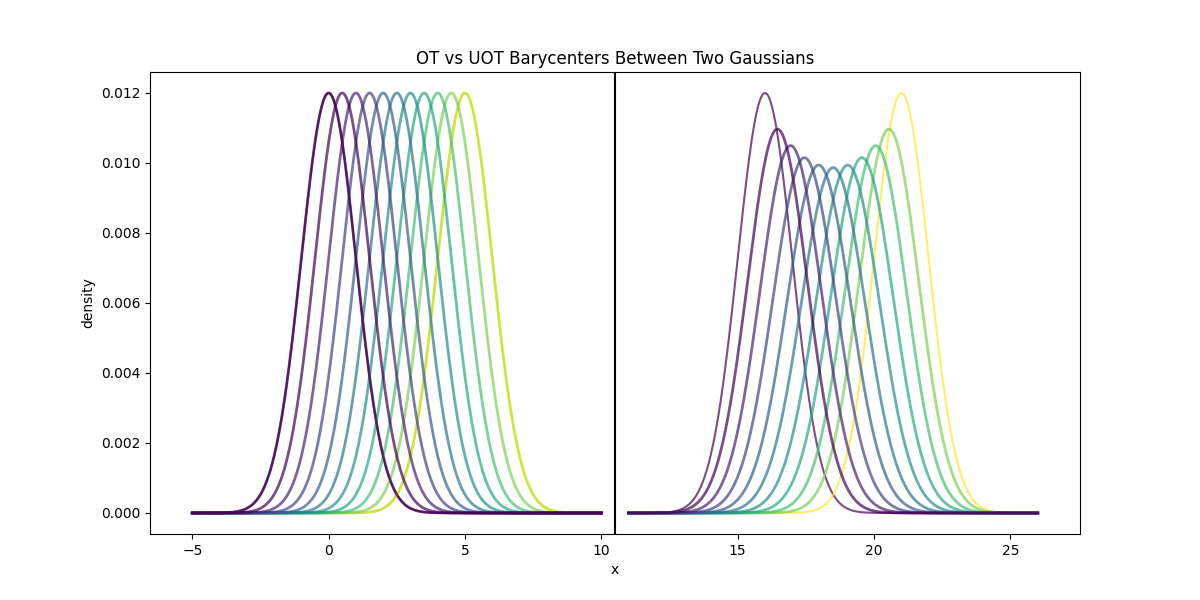}
    \caption{Pictured here is the balanced (left) and unbalanced (right) barycentric interpolation between two Gaussian distributions with the same variance using $\tau = 0.5$ and $\epsilon = 0.001$. Notice that the unbalanced barycenters do not obey strict mass conservation, but still take the general shape of the reference distributions. }
    \label{fig:placeholder}
\end{figure}
%This sentence can be better. WDL is too cool to explain it like this
\indent We can invert the barycenter problem by seeking a set of weights and distributions whose barycenter yields an approximation of a predefined distribution \cite{schmitz2018wasserstein}. Given HSI data $\{\mu_j \}_{j=1}^m \in \mathbbm{R}^d$, we interpret each pixel as a distribution supported on the spectral bands captured by the sensor. We seek to find $k$ dictionary atoms $\{\nu_i\}_{i=1}^k = D \in \mathbbm{R}^{d \times k}_{\geq 0}$, $\nu_i = (\nu_{i,1}, ..., \nu_{i,d})$, and $m$ weight vectors $\Lambda = \{\Lambda_j \}_{j=1}^m \in \mathbbm{R}^{k \times m}$, $\Lambda_j = (\lambda_{j, 1}, ... , \lambda_{j,k})$, such that each barycenter $P(D, \Lambda_i) \approx \mu_i$. Specifically, for some loss function $\mathcal{L}$, we seek to minimize:
\begin{equation}
\begin{aligned}
    \min_{D, \Lambda} \mathcal{E}(D, \Lambda) = \sum_{j=1}^m \mathcal{L}(P(D, \Lambda_i), \mu_i)
\end{aligned}
\end{equation}
The aim of this scheme is to faithfully reconstruct the underlying distributions using a learned set of dictionary atoms and weights. In practice, this enables us to leverage the underlying geometry of large data sets to reduce the scale of the data considerably in the form of the weight matrix $\Lambda$. To solve this non-convex optimization problem, we use an iterative method and automatic gradient approximations to repeatedly update $D$ and $\Lambda$ \cite{schmitz2018wasserstein}. \\

%\indent In the next section we provide a detailed algorithm for solving (4), and detail how we use the learned weight matrix for spectral clustering. 
\section{Clustering HSI with Unbalanced Barycenters}
\label{sec:pagestyle}
In the case of HSI data, we view pixels, $\{\mu_j\}_{j=1}^m$, as distributions supported on the set of reflectance bands. For instance, the $7832$ pixels of the Salinas A data set are supported on $201$ bands spaced roughly $4$nm apart (some are slightly further after removal of water reflectance bands). This framing allows us to generate a reconstruction of the HSI data as a set of learned barycenters. In our case, the cost matrix is a square matrix using quadratic cost between the reflectance bands on which the pixels are supported. In this way, if learned weights $\Lambda_i$ and $\Lambda_j$ are close then the underlying pixels, $\mu_i$ and $\mu_j$, must also have similar distributions. Notably, this was not necessarily true for when using balanced optimal transport and normalizing the data. One could imagine two pixels whose distribution of reflectances differed by a scalar multiple. In this case, the reweighting of distributions to be probabilities would have obscured this difference. By incorporating the information about total reflectance, we expect to get weights which better capture the underlying data. 
%fix above sentence.
\\ 
\indent Our clustering process proceeds in two steps. Firstly, we need to perform the learning phase to generate weights, $\{\Lambda_j\}_{j=1}^m$ of the underlying data $\{\mu_j\}_{j=1}^m$ and then, secondly, we need to perform spectral clustering on the resultant matrix of weights $\Lambda$. In our case we use the metric $d(\mu_i, \mu_j) = ||\Lambda_i - \Lambda_j||^2$ for spectral clustering. \\
\indent The main process for doing this dictionary learning is to (i) initialize a random subset of the underlying data, $X = \{\mu_i \}_{i=1}^n$, (ii) randomly initialize a vector of weights $\Lambda_i$ for each $\mu_i$ and starting dictionary atoms $D_i$, (iii) compute the unbalanced barycenters $P(D, \Lambda_i)$ for all $i$, (iv) compute the loss $\mathcal{L}(P(D, \Lambda_i), \mu_i)$, (v) use automatic differentiation of the loss function to approximate gradients for $D$ and $\Lambda$ and then use backward propagation to update $D$ and $\Lambda$ \cite{paszke2017automatic}. We note it is necessary to impose a floor on the values for $D$ so as the values do not become negative and to normalize the $\Lambda_i$'s to sum to $1$ after this update. For our implementation, we set the non-positive values of $X$ to $1\mathrm{e}{-15}$, a value just above the machine precision limit. This is detailed in Algorithm 1. \\

%Update notation to be in line with above. 
%still don't know if we're going with the name unbalanced Wasserstein dictionary learning. Unbalanced Wasserstein is sort of just not right
\begin{algorithm}[H]
    \caption{Unbalanced Optimal Transport Dictionary Learning}
    \begin{algorithmic}[1]
    \State Initialize $X$, C, $\epsilon$, $\tau$, and initial weights ($\Lambda$)
    \State Initialize optimizer (e.g., ADAM) for updating dictionary atoms $D$ and weights $\Lambda$
    \While{not converged}
        \State $\Lambda \gets \text{Softmax}(\Lambda)$
        \State $P \gets P(D, \Lambda)$ \Comment{Compute barycentric reconstructions of each histogram $x_i$}
        \State $\text{loss} \gets \mathcal{L}(P, X)$ \Comment{Compute loss}
        \State Compute gradients of $D$ and $\Lambda$: loss$.\text{backward()}$
        \State Use gradients to update $D$ and $\Lambda$: $\text{optimizer.step()}$
        \State Set all non-positive entries of $D$ to be  $1\mathrm{e}{-15}$
    \EndWhile
    \State \Return $D$ and $\Lambda$
    \end{algorithmic}
    \end{algorithm}

After learning barycentric weights $\{\Lambda_i \}_{i=1}^n$, they can be used for spectral clustering. We employ the same overall algorithm used in \cite{fullenbaum2024hyperspectral}, barycentric coding spectral clustering, which is detailed in Algorithm 2, unbalanced barycentric coding spectral clustering (UBCSC). It should be noted that, since UBCSC produces labels not inherently aligned with the ground truth classes, we perform an optimal matching step prior to evaluation. Specifically, we apply the Hungarian algorithm to find the permutation of cluster labels that maximizes agreement with the ground truth. All evaluation metrics discused later occur after this matching.

%BSSC 
\begin{algorithm}[H]
    \caption{Unbalanced Barycentric Coding Spectral Clustering (UBCSC)}
    \begin{algorithmic}[1]
        \State \textbf{Inputs:} HSI data: $\{\mu_j \}_{j=1}^m$; number of atoms: $k$; number of nearest neighbors: $N N$; number of clusters: $K$; number of pixels: $n$. 
        \State Select a random subset of $n$ pixels of $\{\mu_j\}_{j=1}^m$, $\{\tilde\mu_i\}_{i=1}^n$
        \State Run algorithm 1 on $\{\tilde\mu_i\}_{i=1}^n$ to learn barycentric weights $\{\Lambda_i\}_{i=1}^n$
        \State Run $K$-means on the $K$ lowest frequency eigenvectors of the symmetric normalize Laplacian of the $N N$ graph associated with weights $\{\Lambda_i\}_{i=1}^n$ to acquire labels $\{\tilde y_i\}_{i=1}^n$. 
        \State Match cluster labels $\{\tilde y_i\}_{i=1}^n$ to ground truth via Hungarian algorithm. 
        \State Assign labels to the $m -n$ unlabeled pixels via majority vote of the $10$ $\ell^1$ nearest neighbors of the labeled pixels $\{\tilde \mu_i\}_{i=1}^n$. These resulting in-painted labels are $\{y_j\}_{j=1}^m$, the fully labeled data set.
        \State Return labels $\{y_j\}_{j=1}^m$
        
    \end{algorithmic}
\end{algorithm}

\indent One important, but as of yet undiscussed, part of the above algorithm is the choice of loss function, $\mathcal{L}(P, X)$. Following the suggestions outlined in \cite{schmitz2018wasserstein} we tried four different loss functions: $TV$, quadratic, $KL$, and a unbalanced transport based loss. Similar to what was observed in \cite{schmitz2018wasserstein}, we found that all choices of loss function perform similarly in creating accurate reconstructions of the underlying data. Notably, the unbalanced transport loss incurs a higher computational cost and does not make up for this deficit in additional accuracy. For all of the tests we ran, we used the quadratic loss given by $\mathcal{L}(P, X) = ||P-X||_2^2 = \sum_{i=1}^n\sum_{j=1}^d (P_{ij} - X_{ij})^2$. We found this resulted in generally faithful reconstructions of the underlying distributions, while maintaining simplicity and run-time efficiency. 

Python code, in part utilizing the POT library \cite{flamary2021pot}, and all experiments are freely available through GitHub at https://github.com/jlentz02/WDL. 
\section{EXPERIMENTAL RESULTS}
\label{sec:typestyle}
We tested unbalanced barycentric coding spectral clustering (UBCSC) on the Salinas A, Indian Pines, Pavia Centre, and Pavia University hyperspectral data sets which are freely available online \cite{grana2021hyperspectral}. We considered two scenarios and have two different metrics for gauging them. In the first scenario we suppose that we know the number of ground truth classes and assign an equal number of labels. In this case, we gauge the performance of the algorithm with labeling accuracy. This is the total number of correctly labeled pixels divided by the total number of pixels before in-painting. In the second scenario, we let the number of labels exceed the number of ground truth classes and score a purity metric based on how well each labeled class represents its majority ground truth. Understandably, the later generates more appealing results, but is mostly for identifying latent classes that are present in the data, but not in the ground truth labeling. For instance, the lower right hand corner of the Salinas A data set is notorious for being mislabeled under unsupervised clustering approaches, with this modification identifying it as two separate classes. \\
\indent Further complicating our approach is the large number of hyperparameters, and their varying effects on the resulting labeling accuracy. Due to computational considerations, we divide each HSI pixel by the total mass of the pixel with highest mass. This results in the total mass of each pixel falling between $0$ and $1$. Similarly, we rescale the cost matrix to have largest value $10$. This rescaling has no effect on the computation of unbalanced barycenters when considering that the hyperparameter $\tau$, the marginal relaxation term, already acts as a scalar on the cost matrix. However, our observed values of $\tau$, which yield strong results, are impacted by this. In general, we observed strong results when $\tau$ is approximately equal to the total mass of the underlying data, $\epsilon$ is in the range $[0.07, 0.12]$, and the number of atoms, $k$, is between two and four times the number of ground truth labels. In general, the values for the number of nearest neighbors, $N N$, is highly subject to the underlying scene. We observed that dense clusterings of similar pixels benefited from higher values, such as in Salinas A, and sparser scenes, such as Pavia Centre, benefited from fewer nearest neighbors. Optimal tuning of hyperparameters, when accounting for effects on run-time and accuracy, is an ongoing subject of investigation, but appears very specific to individual scenes. For this paper, we ran a parameter sweep on all four data sets using $\tau \in [100,1000,10000,100000]$, $\epsilon \in [0.05,0.06,0.07,0.08,0.09,0.1]$, and $NN \in [5,10,...,45,50]$. The number of atoms used was a multiple, $0.5, 1, 2, \text{or } 4$, of the number of ground truth labels which vary for each data set. When we express best case performance, we mean from the results returned by this sweep. \\
\indent The following table indicates best case performance accuracy under a few different sets of hyperparameters for our four trial data sets.

\begin{table}[h!]
\centering
\begin{tabular}{l c c c c c}
\hline
\textbf{Data set} & \textbf{Accuracy} & $\boldsymbol{\tau}$ & $\boldsymbol{\epsilon}$ & \textbf{Atoms} & \textbf{NN} \\
\hline
Salinas A & 0.89 & 1000 & 0.1 & 24 & 25 \\
Salinas A & 0.86 & 10000 & 0.05 & 30 & 20 \\
Pavia Centre & 0.84 & 1000 & 0.07 & 27 & 15 \\
Pavia Centre & 0.82 & 100 & 0.07 & 27 & 5 \\
Pavia U & 0.63 & 10000 & 0.07 & 18 & 5 \\
Pavia U & 0.58 & 100000 & 0.1 & 36 & 5 \\
Indian Pines & 0.34 & 1000 & 0.06 & 32 & 5 \\
Indian Pines & 0.34 & 100000 & 0.06 & 48 & 5 \\
\hline
\end{tabular}
\caption{Accuracy test results for various data sets.}
\label{tab:results}
\end{table}

%purity data and discussion
Often, when presented with raw HSI data, it is unclear what the number of ground truth classes should be. This motivates the purity metric of evaluating a labeling scheme. In this regime, we do not presuppose the number of ground truth classes, and instead cluster around $c$ clusters. For each of the $c$ clusters, we find the most common ground truth label within the cluster. We then compute the ratio of number of pixels sharing this consensus label to the total number of pixels within the cluster. The average of these ratios is the purity score. It captures how effectively each cluster parses ground truth classes, even if the number of clusters exceeds the number of ground truth classes. \\
\indent While the purity score is still presented as a value between $0$ and $1$, there is nuance required when comparing it to accuracy. For instance, the purity score is subject to the number of clusters, $c$, with purity approaching $1$ as $c \to \infty$. The following table presents purity test results from our four data sets, highlighting the advantages of adding a few additional clusters, rather than the best case outcome which is overly appealing for large $c$. In particular, we note the addition of even one additional cluster to the Salinas A data set results in an improvement from an accuracy of $89\%$ to a purity of $92\%$ albeit with different hyperparameters. Similar results for the improvement of purity over accuracy with one additional cluster point, in our opinion, to the existence of latent material classes within these data sets. In table 2, $c$ is the number of clusters. For reference the number of ground truth classes for Salinas A, Pavia Centre, Pavia University, and Indian Pines are 6, 9, 9, and 16 respectively. 
\setlength{\tabcolsep}{4pt}
\begin{table}[h!]
\centering
\begin{tabular}{l c c c c c c c}
\hline
\textbf{Data set} & \textbf{Purity} & $\boldsymbol{\tau}$ & $\boldsymbol{\epsilon}$ & \textbf{Atoms} & \textbf{NN} & \textbf{c} \\
\hline
Salinas A & 0.92 & 1000 & 0.1 & 60 & 45 & 7 \\
Pavia Centre & 0.91 & 10000 & 0.09 & 24 & 10 & 10 \\
Pavia U & 0.72 & 10000 & 0.11 & 36 & 20 & 10 \\
Indian Pines & 0.47 & 1000 & 0.1 & 64 & 40 & 17 \\ 
Salinas A & 0.93 & 1000 & 0.1 & 30 & 45 & 35 \\
Pavia Centre & 0.91 & 10000 & 0.09 & 24 & 10 & 18 \\
Pavia U & 0.75 & 10000 & 0.11 & 36 & 10 & 13 \\
Indian Pines & 0.53 & 1000 & 0.1 & 64 & 25 & 44 \\ 

\hline
\end{tabular}
\caption{Purity test results for various data sets. The first four rows highlight purity results in which the number of clusters exceeds the number of ground truth labels by just one. The following four rows allow $c$ to further exceed the number of ground truth labels. }
\label{tab:results}
\end{table}
\setlength{\tabcolsep}{6pt}

\begin{figure}
    \centering
    \includegraphics[width=1\linewidth]{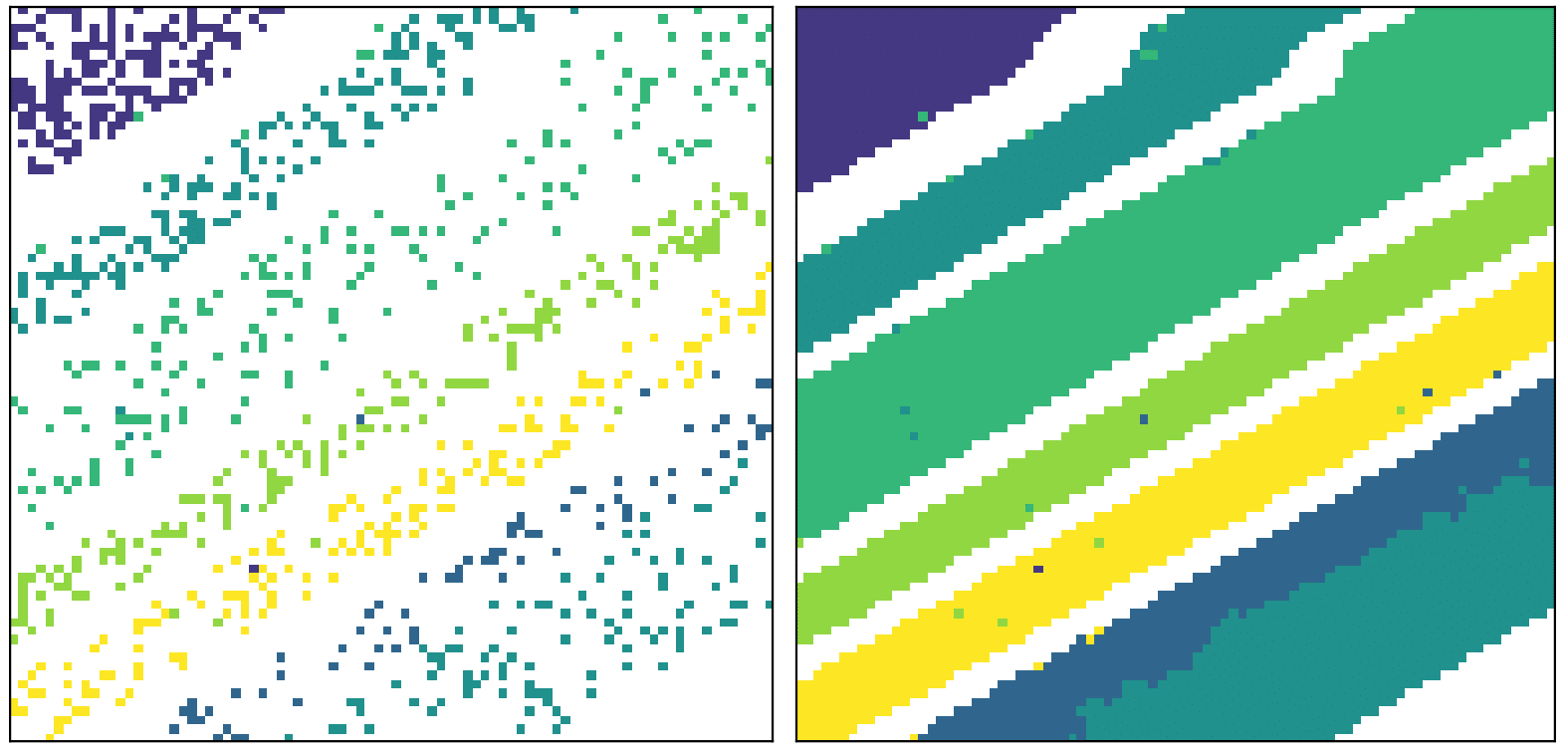}
    \caption{This image shows the in-painting process of a trial run of the Salinas A data set with $24$ atoms, $\tau = 1000$, and $\epsilon = 0.1$ achieving an accuracy before in-painting of $89\%$. We note that the bottom right corner of the image accounts for the majority of the mislabeling, and that this issue is common under unsupervised labeling schemes.}
    \label{fig:placeholder}
\end{figure}

%comparison to UBCSC/BCSC
Much of the work done in this paper sought to improve upon the results showcased in \cite{fullenbaum2024hyperspectral} which deployed BCSC as opposed to UBCSC. In the following tables we highlight the performance difference between UBCSC and BCSC in terms of labeling accuracy. The first table showcases trials under identical hyperparameters, and the second table the overall best case performance on each data set between the two algorithms.

\begin{table}[h!]
\centering
\begin{tabular}{l c c c c c}
\hline
\textbf{Data set} & \textbf{UBCSC} & \textbf{BCSC} & $\boldsymbol{\tau}$ & $\boldsymbol{\epsilon}$ & \textbf{Atoms} \\
\hline
Salinas A & 0.89 & 0.68 & 1000 & 0.1 & 24 \\
Pavia Centre & 0.84 & 0.73 & 1000 & 0.07 & 27 \\
Pavia U & 0.63 & 0.40 & 10000 & 0.07 & 18 \\
Indian Pines & 0.34 & 0.30 & 1000 & 0.06 & 32 \\
\hline
\end{tabular}
\caption{Comparison of UBCSC vs BCSC accuracy under identical hyperparameters.}
\label{tab:results}
\end{table}

\begin{table}[h!]
\centering
\begin{tabular}{l c c }
\hline
\textbf{Data set} & \textbf{UBCSC} & \textbf{BCSC} \\
\hline
Salinas A & 0.89 & 0.86  \\
Pavia Centre & 0.84 & 0.77  \\
Pavia U & 0.63 & 0.47  \\
Indian Pines & 0.34 & 0.32  \\
\hline
\end{tabular}
\caption{Comparison of UBCSC vs BCSC accuracy under best case performance.}
\label{tab:results}
\end{table}

To conclude this section, we note that one limitation with our approach its time complexity. Unlike one dimensional balanced optimal transport where we can compute a barycenter on $n$ distributions in $O(n\log(n))$ time via a sorting algorithm, no such efficient solution exists for one dimensional unbalanced optimal transport. However, the prior implementation of BCSC did not employ this approach and instead used Sinkhorn’s algorithm, as does our implementation of UBCSC. It is well known that Sinkhorn’s algorithm has time complexity $O(n^2/\epsilon^2)$ for balanced optimal transport and $O(n^2/\epsilon)$ for unbalanced optimal transport \cite{altschuler2017nearlinear,pham2020unbalanced}. Despite this, we have observed that UBCSC is slower than BCSC under similar hyperparameters. Moreover, the quadratic time complexity would render this method untenable for large data sets ($n>10000$). For our experiments, we utilized a system running Windows 11 with a Ryzen 9700x CPU (8 cores, 16 threads, 3.8-5.5GHz) and 32GB of memory  running single threaded in Python 3.14. It is possible to parallelize the barycenter computation on a GPU to expedite computations, but this is not something we have implemented. As an example, our best case scenario on the Salinas A data set took 226 seconds of runtime with hyperparameters: $k = 24, \tau = 1000, \epsilon = 0.1$, and 500 iterations of the learning loop.

\section{CONCLUSIONS AND FUTURE WORK}
\label{sec:majhead}
In this paper we demonstrated the efficacy of applying unbalanced optimal transport barycenters towards creating a learned representation of HSI data. Unbalanced Wasserstein dictionary learning meaningfully reduced the dimensionality of the problem which, in conjunction with standard clustering techniques, provided an effective unsupervised labeling scheme. We have largely improved on the prior work of \cite{fullenbaum2024hyperspectral}, but have done so in a way that is both more faithful to the underlying HSI data and more effective.  \\
\indent For future work, we seek to incorporate additional spatial data in the labeling process, either when generating the weights, or as a post-processing step. One risk of incorporating additional spatial information is overfitting as certain data sets feature spatial separated areas of similar material type. Regardless, it is an unused component of the data which could offer further benefit if incorporated.

% References should be produced using the bibtex program from suitable
% BiBTeX files (here: strings, refs, manuals). The IEEEbib.bst bibliography
% style file from IEEE produces unsorted bibliography list.
% -------------------------------------------------------------------------
\bibliographystyle{IEEEbib}
\bibliography{strings,refs}

@article{ahmad2025comprehensive,
  title={A comprehensive survey for hyperspectral image classification: The evolution from conventional to transformers and mamba models},
  author={Ahmad, Muhammad and Distefano, Salvatore and Khan, Adil Mehmood and Mazzara, Manuel and Li, Chenyu and Li, Hao and Aryal, Jagannath and Ding, Yao and Vivone, Gemine and Hong, Danfeng},
  journal={Neurocomputing},
  pages={130428},
  year={2025},
  publisher={Elsevier}
}

@article{kumar2024deep,
  title={Deep learning for hyperspectral image classification: A survey},
  author={Kumar, Vinod and Singh, Ravi Shankar and Rambabu, Medara and Dua, Yaman},
  journal={Computer Science Review},
  volume={53},
  pages={100658},
  year={2024},
  publisher={Elsevier}
}

@article{villani2008optimal,
  author = {Cedric Villani},
  title = {Optimal Transport, old and new},
  journal = {Springer Science and \& Buisness Media},
  year = {2008},
  volume = {338}
}

@article{schmitz2018wasserstein,
  author = {Morgan Schmitz and Matthieu Heitz and Nicolas Bonneel and Fred Ngole and David Coeurjolly and Marco Cuturi and Gabril Peyre and Jean-Luc Starck},
  title = {Wasserstein Dictionary Learning: Optimal Transport-Based Unsupervised Nonlinear Dictionary Learning},
  journal = {SIAM J. Imaging Sciences},
  year = {2018},
  volume = {11},
  number = {1},
  pages = {643-678}
}

@article{chizat2015unbalanced,
  author = {Lenaic Chizat and Gabriel Peyre and Bernhard Schmitzer and Francois-Xavier Villard},
  title = {Unbalanced Optimal Transport: Dynamic and Kantorovich Formulations},
  journal = {Science Direct},
  year = {2015},
  volume = {274},
  number = {11},
  pages = {3090-3123}
}

@article{chizat2017scaling,
      title={Scaling Algorithms for Unbalanced Transport Problems}, 
      author={Lenaic Chizat and Gabriel Peyré and Bernhard Schmitzer and François-Xavier Vialard},
        journal = {Mathematics of Computation},
      year={2017},
      eprint={1607.05816},
pages = {2563–2609},
      archivePrefix={arXiv},
      primaryClass={math.OC},
      url={https://arxiv.org/abs/1607.05816}, 
}

@article{agueh2011barycenters,
author = {Agueh, Martial and Carlier, Guillaume},
title = {Barycenters in the {W}asserstein Space},
journal = {SIAM Journal on Mathematical Analysis},
volume = {43},
number = {2},
pages = {904-924},
year = {2011},
doi = {10.1137/100805741},
URL = { https://doi.org/10.1137/100805741},
eprint = { https://doi.org/10.1137/100805741}
}

@INPROCEEDINGS{fullenbaum2024hyperspectral,
  author={Fullenbaum, Scott and Mueller, Marshall and Tasissa, Abiy and Murphy, James M.},
  booktitle={IGARSS 2024 - 2024 IEEE International Geoscience and Remote Sensing Symposium}, 
  title={Hyperspectral Image Clustering Via Learned Representation In Wasserstein Space}, 
  year={2024},
  volume={},
  number={},
  pages={8791-8796},
  doi={10.1109/IGARSS53475.2024.10641170}
}

@article{melgani2004classifiation,
  author    = {Farid Melgani and Lorenzo Bruzzone},
  title     = {Classification of hyperspectral remote sensing images with support vector machines},
  journal   = {IEEE Transactions on Geoscience and Remote Sensing},
  volume    = {42},
  number    = {8},
  pages     = {1778--1790},
  year      = {2004}
}

@article{ham2005investigation,
  author    = {Jisoo Ham and Yangchi Chen and Melba M. Crawford and Joydeep Ghosh},
  title     = {Investigation of the random forest framework for classification of hyperspectral data},
  journal   = {IEEE Transactions on Geoscience and Remote Sensing},
  volume    = {43},
  number    = {3},
  pages     = {492--501},
  year      = {2005}
}

@article{li2019deep,
  author    = {Shutao Li and Weiwei Song and Leyuan Fang and Yushi Chen and Pedram Ghamisi and Jon Atli Benediktsson},
  title     = {Deep learning for hyperspectral image classification: An overview},
  journal   = {IEEE Transactions on Geoscience and Remote Sensing},
  volume    = {57},
  number    = {9},
  pages     = {6690--6709},
  year      = {2019}
}

@article{zhai2021hyperspectral,
  author    = {Han Zhai and Hongyan Zhang and Pingxiang Li and Liangpei Zhang},
  title     = {Hyperspectral image clustering: Current achievements and future lines},
  journal   = {IEEE Geoscience and Remote Sensing Magazine},
  volume    = {9},
  number    = {4},
  pages     = {35--67},
  year      = {2021}
}

@inproceedings{balaji2020robust,
      title={Robust Optimal Transport with Applications in Generative Modeling and Domain Adaptation}, 
      author={Yogesh Balaji and Rama Chellappa and Soheil Feizi},
booktitle = {NeurIPS},
      year={2020},
pages = {12934 - 12944},
      eprint={2010.05862},
      archivePrefix={arXiv},
      primaryClass={cs.LG},
      url={https://arxiv.org/abs/2010.05862}, 
}

@article{shi2000normalized,
  author    = {Jianbo Shi and Jitendra Malik},
  title     = {Normalized cuts and image segmentation},
  journal   = {IEEE Transactions on Pattern Analysis and Machine Intelligence},
  volume    = {22},
  number    = {8},
  pages     = {888--905},
  year      = {2000}
}

@inproceedings{ng2001spectral,
  author    = {Andrew Ng and Michael Jordan and Yair Weiss},
  title     = {On spectral clustering: Analysis and an algorithm},
  booktitle = {Advances in Neural Information Processing Systems (NeurIPS)},
pages = {849 - 856},
  volume    = {14},
  year      = {2001}
}

@inproceedings{cahill2014schrodinger,
  author    = {Nathan D. Cahill and Wojciech Czaja and David W. Messinger},
  title     = {Schr{\"o}dinger eigenmaps with nondiagonal potentials for spatial-spectral clustering of hyperspectral imagery},
  booktitle = {Algorithms and Technologies for Multispectral, Hyperspectral, and Ultraspectral Imagery XX},
  series    = {SPIE Proceedings},
  volume    = {9088},
  pages     = {27--39},
  year      = {2014}
}

@article{vonLuxburg2007tutorial,
  author    = {Ulrike von Luxburg},
  title     = {A tutorial on spectral clustering},
  journal   = {Statistics and Computing},
  volume    = {17},
  pages     = {395--416},
  year      = {2007}
}

@inproceedings{cuturi2013sinkhorn,
      title={Sinkhorn Distances: Lightspeed Computation of Optimal Transportation Distances}, 
booktitle = {International Conference on Neural Information Processing Systems},
      author={Marco Cuturi},
      year={2013},
      eprint={1306.0895},
pages = {2292 - 2300},
      archivePrefix={arXiv},
      primaryClass={stat.ML},
      url={https://arxiv.org/abs/1306.0895}, 
}

@inproceedings{paszke2017automatic,
  title={Automatic differentiation in PyTorch},
  author={Paszke, Adam and Gross, Sam and Chintala, Soumith and Chanan, Gregory and Yang, Edward and DeVito, Zachary and Lin, Zeming and Desmaison, Alban and Antiga, Luca and Lerer, Adam},
  booktitle={NeurIPS Autodiff Workshop},
  year={2017}
}

@article{flamary2021pot,
  author  = {R{\'e}mi Flamary and Nicolas Courty and Alexandre Gramfort and Mokhtar Z. Alaya and Aur{\'e}lie Boisbunon and Stanislas Chambon and Laetitia Chapel and Adrien Corenflos and Kilian Fatras and Nemo Fournier and L{\'e}o Gautheron and Nathalie T.H. Gayraud and Hicham Janati and Alain Rakotomamonjy and Ievgen Redko and Antoine Rolet and Antony Schutz and Vivien Seguy and Danica J. Sutherland and Romain Tavenard and Alexander Tong and Titouan Vayer},
  title   = {POT: Python Optimal Transport},
  journal = {Journal of Machine Learning Research},
  year    = {2021},
  volume  = {22},
  number  = {78},
  pages   = {1-8},
  url     = {http://jmlr.org/papers/v22/20-451.html}
}

@misc{grana2021hyperspectral,
  author       = {Graña, M. and Veganzons, M.~A. and Ayerdi, B.},
  title        = {Hyperspectral Remote Sensing Scenes},
  howpublished = {Grupo de Inteligencia Computacional (GIC), University of the Basque Country},
  year         = {2021},
  note         = {Last edited 12 July 2021. Accessed: 15 September 2025},
  url          = {https://www.ehu.eus/ccwintco/index.php/Hyperspectral_Remote_Sensing_Scenes}
}

@INPROCEEDINGS{fullenbaum2024nonlinear,
  author={Fullenbaum, Scott and Mueller, Marshall and Tasissa, Abiy and Murphy, James M.},
  booktitle={IGARSS 2024 - 2024 IEEE International Geoscience and Remote Sensing Symposium}, 
  title={Nonlinear Unmixing of Hyperspectral Images via Regularized Wasserstein Dictionary Learning}, 
  year={2024},
  volume={},
  number={},
  pages={8289-8294},
  doi={10.1109/IGARSS53475.2024.10642450}}

@inproceedings{mueller2023geometrically,
  title={Geometrically regularized wasserstein dictionary learning},
  author={Mueller, Marshall and Aeron, Shuchin and Murphy, James M and Tasissa, Abiy},
  booktitle={Topological, Algebraic and Geometric Learning Workshops 2023},
  pages={384--403},
  year={2023},
  organization={PMLR}
}

@inproceedings{pham2020unbalanced,
      title={On Unbalanced Optimal Transport: An Analysis of {S}inkhorn Algorithm}, 
      author={Khiem Pham and Khang Le and Nhat Ho and Tung Pham and Hung Bui},
        booktitle = {International Conference on Machine Learning},
      year={2020},
pages = {7673 - 7682},
      eprint={2002.03293},
      archivePrefix={arXiv},
      primaryClass={cs.CC},
      url={https://arxiv.org/abs/2002.03293}, 
}

@inproceedings{altschuler2017nearlinear,
      title={Near-linear time approximation algorithms for optimal transport via {S}inkhorn iteration}, 
        booktitle = {Advances in Neural Information Processing Systems},
      author={Jason Altschuler and Jonathan Weed and Philippe Rigollet},
      year={2017},
pages = {1961 - 1971},
      eprint={1705.09634},
      archivePrefix={arXiv},
      primaryClass={cs.DS},
      url={https://arxiv.org/abs/1705.09634}, 
}

\end{document}